\documentclass[letterpaper]{article} % DO NOT CHANGE THIS
\usepackage{aaai25}  % DO NOT CHANGE THIS
\usepackage{times}  % DO NOT CHANGE THIS
\usepackage{helvet}  % DO NOT CHANGE THIS
\usepackage{courier}  % DO NOT CHANGE THIS
\usepackage[hyphens]{url}  % DO NOT CHANGE THIS
\usepackage{graphicx} % DO NOT CHANGE THIS
\usepackage{natbib}  % DO NOT CHANGE THIS AND DO NOT ADD ANY OPTIONS TO IT
\usepackage{caption} % DO NOT CHANGE THIS AND DO NOT ADD ANY OPTIONS TO IT
\usepackage{algorithm}
\usepackage{algorithmic}
\usepackage{amsmath} 
\usepackage{amssymb}
\usepackage{booktabs}
\usepackage{multirow}
\usepackage{pifont}
\usepackage{xcolor}
\usepackage{newfloat}
\usepackage{listings}

\DeclareCaptionStyle{ruled}{labelfont=normalfont,labelsep=colon,strut=off} % DO NOT CHANGE THIS
\floatstyle{ruled}
\newfloat{listing}{tb}{lst}{}
\floatname{listing}{Listing}
\title{MVREC: A General Few-shot Defect Classification Model Using Multi-View Region-Context}
\author{
    Shuai Lyu\textsuperscript{\rm 1 \rm 2}, Rongchen Zhang\textsuperscript{\rm 1 \rm 2}, Zeqi Ma\textsuperscript{\rm 1 \rm 2}\\
    Fangjian Liao\textsuperscript{\rm 1 \rm 2}, Dongmei Mo\textsuperscript{\rm 1 \rm 2}, Waikeung Wong \textsuperscript{\rm 1 \rm 2}\thanks{Corresponding author.} \\
}
\affiliations{
    \textsuperscript{\rm 1}  The Hong Kong Polytechnic University, Hong Kong SAR, China\\
    \textsuperscript{\rm 2} Laboratory for Artificial Intelligence in Design, Hong Kong SAR, China\\
    \{shuai.lyu, rongchen.zhang, mzeqi.ma, fangjian.liao, dongmei.mo\}@connect.polyu.hk \\
    calvin.wong@polyu.edu.hk
}
\usepackage{bibentry}
\begin{document}

\maketitle

\begin{abstract}

Few-shot defect multi-classification (FSDMC) is an emerging trend in quality control within industrial manufacturing. However, current FSDMC research often lacks generalizability due to its focus on specific datasets. Additionally, defect classification heavily relies on contextual information within images, and existing methods fall short of effectively extracting this information. To address these challenges, we propose a general FSDMC framework called MVREC, which offers two primary advantages: (1) MVREC extracts general features for defect instances by incorporating the pre-trained AlphaCLIP model. (2) It utilizes a region-context framework to enhance defect features by leveraging mask region input and multi-view context augmentation. Furthermore, Few-shot Zip-Adapter(-F) classifiers within the model are introduced to cache the visual features of the support set and perform few-shot classification. We also introduce MVTec-FS, a new FSDMC benchmark based on MVTec AD, which includes 1228 defect images with instance-level mask annotations and 46 defect types. Extensive experiments conducted on MVTec-FS and four additional datasets demonstrate its effectiveness in general defect classification and its ability to incorporate contextual information to improve classification performance. 
%Extensive experiments conducted on MVTec-FS and four additional datasets demonstrate the effectiveness of MVREC.

\end{abstract}
\begin{links}
    \link{Code}{https://github.com/ShuaiLYU/MVREC}
\end{links}

\section{Introduction}

Defect detection and classification~\cite{dd_review} is a critical challenge in industrial manufacturing, as it involves identifying and categorizing defects within workpieces. However, in practical applications, the diversity of defect types and the low frequency of defect occurrences make it a particularly difficult task.

\begin{figure}[t]
    \centering  % 更简洁的居中命令，不会增加额外的垂直空间
    \includegraphics[width=3.3in]{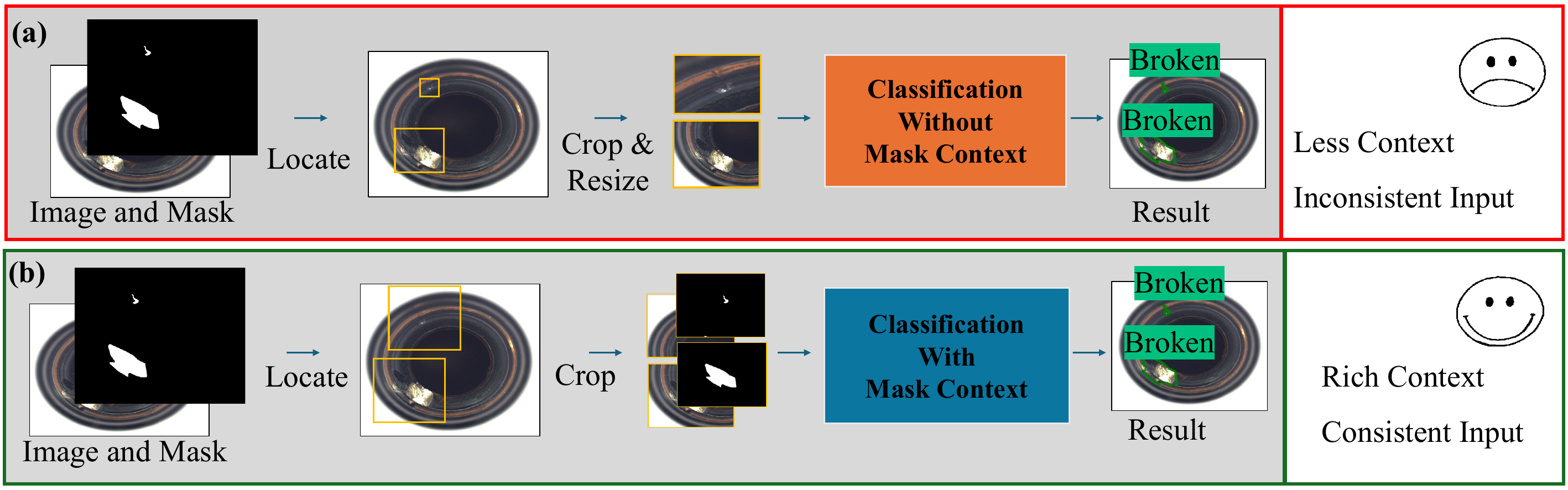}  % 确保图片文件名和路径正确
    \caption{Comparison of two different Classification models.}
    \label{head_pic}  % 为图像添加标签，便于文中引用
\end{figure}

% gap 1 
While Few-shot Learning (FSL)~\cite{fs_review} has gained traction in general vision tasks like mini-Imagenet, its application to defect multi-classification (FSDMC) remains challenging. This disparity is evident in the limited availability of dedicated datasets and research focusing on FSDMC. Although Contrastive Vision-Language Pre-training (CLIP)~\cite{clip} has demonstrated remarkable success in learning visual features from large-scale image-text pairs and adapting to downstream tasks with few-shot learning, this type of application is nearly absent in the context of FSDMC. This is primarily due to the significant domain gap between general vision tasks and FSDMC.
% Gap 2 
Secondly, defects inherently differ from normal surface areas, necessitating more contextual information for effective detection and classification. However, common classification models often involve cropping the defect region, resizing it to the model input size, and feeding it into a network, as shown in Figure~\ref{head_pic} (a). This pretreatment fails to retain important contextual information, such as the surrounding background and the size of the defect.
% Gap dataset
The most popular multi-category datasets~\cite{MVTEC,mvtec_loco,visa_data} with different product images are typically designed for anomaly detection rather than defect classification. Although the field of few-shot defect multi-classification has attracted considerable research attention~\cite{CAO2023107294,zhan2022fabric,zhao2023fanet,zhou2023few,liu2023few,xiao2022graph}, the datasets used, such as the NEU-DET Dataset and the MTD Dataset, are limited by their focus on a single product category. There is a notable scarcity of multi-category datasets specifically proposed for FSDMC.

%  motivation (aim )
To mitigate these issues, we propose a general few-shot defect classification model using a multi-view region-context approach, called MVREC. Specifically, our approach begins by generating region-context visual features for the defect instance using the AlphaCLIP model~\cite{alpha_clip}, a transformer-based model that takes a defect image and its mask context as input to generate visual features from the masked region. By incorporating the mask region context, the network can perceive both the defect foreground region and its surrounding background, generating target-specific features while maintaining input consistency. Furthermore, we propose a multi-view augmentation technique to generate multi-view features for a defect, maximizing the utility of few-shot samples and enhancing generalization ability. The multi-view region-context (MVREC) features can be extracted from the multi-view patches and masks of the defect instance, thereby enhancing the region-context features. Moreover, we propose two few-shot classifiers: the training-free Zip-Adapter, which predicts directly without training, and the fine-tuning Zip-Adapter-F, which adapts the MVREC features for better performance. Zip-Adapter and Zip-Adapter-F share the same structure, consisting of a Zero-initialized Projection (ZIP) module and a Scale-Dot-Product Attention (SDPA) module. Specifically, they store visual features and corresponding class labels from the support set images as key-value pairs. The SDPA module then calculates the visual feature similarity between the query defect instance and the support defects, outputting the classification logits through the weighted sum of the encoded labels from the support set. The ZIP module serves as an identity mapping and feature adapter, respectively, for Zip-Adapter and Zip-Adapter-F. Furthermore, we propose MVTec-FS, based on MVTec AD, to create a multi-category dataset suitable for the FSDMC task. This dataset features a diverse array of defect types and a balanced distribution. MVTec-FS includes 15 categories of product surface images and approximately 46 types of defects, making it a promising new benchmark in this field.

We tested MVREC across MVTec-FS and four other public defect datasets with classification annotations. Our results demonstrate superior performance in few-shot defect classification, outperforming existing models. 
In summary, our contributions can be outlined as follows: \\
\textbf{(1)} 
We employ AlphaCLIP to extract general features from each defect instance, enhancing model generalizability, and design a new region-context-based defect classification framework that fully incorporates contextual information for more accurate defect classification.
\\\textbf{(2)} We introduce multi-view context augmentation and Zip-Adapter(-F) classifiers for few-shot classification. \\\textbf{(3)} We reconstruct the popular MVTec AD dataset into a new FSDMC benchmark named MVTec-FS. \\\textbf{(4)} We conducted extensive experiments on multiple defect datasets, demonstrating the effectiveness of MVREC.
 \section{Related work}
 \subsection{Defect Detection and Classification}
Various models, including object detection, segmentation, and classification, have been applied to defect detection. In recent years, the MVTec AD~\cite{MVTEC} dataset has been widely studied for anomaly detection tasks~\cite{REB,ad_review}. These models learn from normal samples to identify anomalies. However, defect classification, which involves identifying specific defect types, is more challenging due to the rarity and diversity of defects and the limited availability of relevant datasets. FSDMC models, such as CAO~\cite{CAO2023107294}, Fabric~\cite{zhan2022fabric}, and FANet~\cite{zhao2023fanet}, have been proposed. However, these methods are often dataset-specific and require complex training processes, including meta-learning and metric learning. Additionally, using a subset of defect types as base classes to train a base model and then evaluating novel classes is common, but this approach may not be practical for real-world applications.
\subsection{Clip-based Few-shot Classification}
The most common few-shot methods include meta-learning and metric learning. Meta-learning methods learn a model that can quickly adapt to new tasks with minimal training data. Metric learning methods learn a distance metric that can effectively measure the similarity between samples. Recently, large language models and multi-modal pre-training models, such as GPT~\cite{gpt3.5, MiniGPT4} and CLIP~\cite{clip}, have emerged as powerful tools. Related research~\cite{AnomalyGPT,winclip} has been applied to defect detection tasks, showing impressive performance. Numerous adapter-based methods have been proposed to adapt the CLIP model to specific tasks with few samples, such as CLIP-Adapter~\cite{clip_adapter}, Tip-Adapter~\cite{tip_adpater}, CoOp~\cite{CoOp}, and SuS-X~\cite{SuS-X}, but most of these methods are designed to jointly learn image and text features for general vision tasks.

\subsection{Region-Context-based Models}
Traditional classification networks typically evolve by cropping the defect region and resizing it, without explicitly utilizing the region context. When defects vary in size, as shown in the example in Figure~\ref{head_pic}, cropping and resizing the defect region may result in the loss of crucial contextual information. To address this issue, region-context models incorporate region context as a prompt to predict target information, thereby preserving the contextual information of the target. For example, the SAM network uses prompts in the form of points and bounding boxes. To enable CLIP to focus on specific regions within the entire image, various methods~\cite{region_clip,dense_label_from_clip,redcircle,alpha_clip} have been explored. AlphaCLIP is an innovative enhancement of the CLIP model, designed to improve its ability to focus attention on specific regions \cite{alpha_clip}. This architecture enables AlphaCLIP to provide precise control over the emphasis of image content.

\begin{figure*}[ht]
    \centering  % 更简洁的居中命令，不会增加额外的垂直空间
    \includegraphics[width=7in]{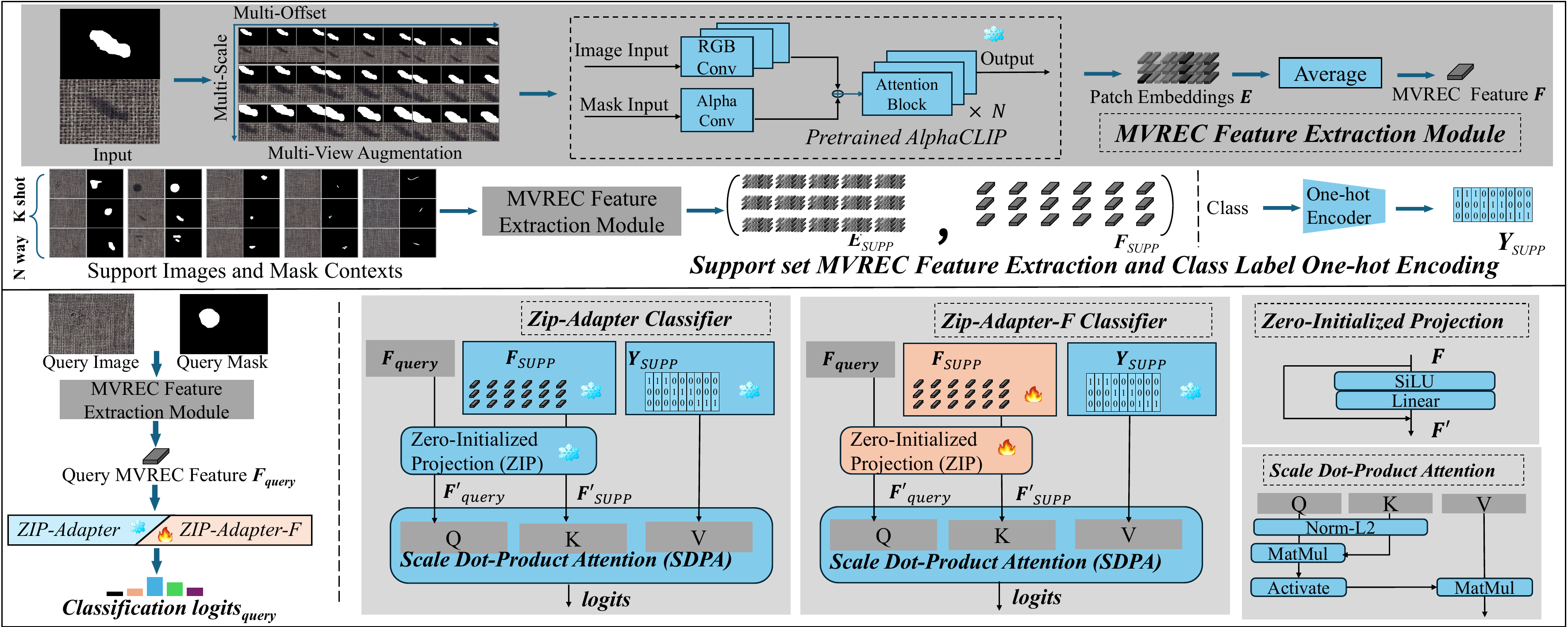}  % 确保图片文件名和路径正确

    \caption{The framework of MVREC. First, the MVREC feature extraction is introduced. Given an N-way K-shot task, the MVREC features for the support set are collected. Then the inference process with Zip-Adapter or Zip-Adapter-F is shown.}  % 添加描述性的标题
    \label{main_pic}  % 为图像添加标签，便于文中引用
\end{figure*}

\section{MVREC}
In this chapter, we introduce the MVREC feature extraction module and present the training-free Zip-Adapter classifier, along with its fine-tuning version, Zip-Adapter-F.

\subsection{Multi-View Region-Context Feature Extraction}
We first introduce the multi-view region-context (MVREC) feature extraction process for defect instances. To effectively capture the visual representation of defects and explicitly mine contextual information, we employ the pre-trained AlphaCLIP model to extract visual features from images using their mask prompts. Given the small data volume characteristic of few-shot learning tasks, we utilize multi-view context augmentation to generate multi-view patches of defect images, thereby expanding the available dataset for subsequent processing. Specifically, we employ two context augmentation methods to achieve this. The first method, multi-scale augmentation, involves cropping $\text{Num}_{scale}$ patches at different scales from the defect patches and their corresponding masks, centered on the defect. The second method involves offsetting the center of the defect to generate $\text{Num}_{offset}$ defect patches with different offsets at each scale. By applying these two augmentation methods, a total of $V = \text{Num}_{scale} \times \text{Num}_{offset}$ patches can be obtained for each defect instance. The AlphaCLIP model extracts MVREC patch embeddings $E \in \mathbb{R}^{V \times C}$ from the multi-view patches, where \(C\) is the number of feature channels. These embeddings are then averaged to produce a single MVREC feature $F \in \mathbb{R}^{C}$.

 \subsection{Support set MVREC Feature Extraction and Class Label Encoding}

For $N$-way $K$-shot classification tasks, the MVREC patch embeddings $E_{SUPP} \in \mathbb{R}^{NK \times V \times C}$ and MVREC features $F_{SUPP} \in \mathbb{R}^{NK \times C}$ are first extracted for the support set. Then, the one-hot encoded class labels $Y_{SUPP} \in \mathbb{R}^{NK \times N}$ are extracted. The MVREC features $F_{SUPP}$ and $Y_{SUPP}$ are used to build the cached key-value pairs for the FSDMC task. Additionally, the MVREC patch embeddings $E_{SUPP}$ and the one-hot encoded labels $Y_{SUPP}$ are used as training data to fine-tune the Zip-Adapter.

\subsection{Training-free Zip-Adapter Classifier} 

In this section, we introduce the method of utilizing MVREC visual features to construct the zero-initialized projection classifier (Zip-Adapter) for FSDMC tasks. The Zip-Adapter classifier consists of a zero-initialized projection (ZIP) module and a scaled dot-product attention (SDPA) module. It stores the MVREC features $F_{SUPP}$ along with the encoded labels $Y_{SUPP}$ of the support set samples. The ZIP module includes a single linear layer, a residual connection, and a SiLU activation function, with the linear layer initialized to zeros. The output of the ZIP module is the adapted feature ${F^{\prime}}$, generated as follows:

   {\small \begin{equation}
        F^{\prime}=\text{SiLU}\left(\text{Linear}\left(F\right)\right) + F,
    \end{equation}}
here, $F$ represents the MVREC feature for either the support sample or the query sample. In the Zip-Adapter, the ZIP module is designed to serve as an identity transformation by initializing the linear layer with zeros and using a residual connection. The SDPA module is a scaled dot-product attention mechanism, which calculates the visual similarity between the query defect instance and the support set. It then produces the classification logits by performing a weighted sum of the support encoded labels $Y_{SUPP}$. The SDPA module is defined as follows:
    % \cdot class_{encoderd}
    {\small\begin{equation}
        logits_{query} = Y_{SUPP} \cdot \psi\left(Sim\left(F^{\prime}_{query}, F^{\prime}_{SUPP} \right)\right),
    \end{equation}}
    where the $\psi$ is the activation function~\cite[]{tip_adpater} for modulating the cosine similarity:
    {\small\begin{equation}
        \psi(x) = \exp\left(-\beta(1 - x)\right),
    \end{equation}}
    $\beta$ controls the sharpness of the curve. And $Sim$ is the cosine similarity function.
    % \begin{equation}
    %     Sim(F^{\prime}_{query}, F^{\prime}_{SUPP}) = \frac{F^{\prime}_{SUPP} F^{\prime}_{query}}{\|F^{\prime}_{SUPP}\| \|F^{\prime}_{query}\|}
    % \end{equation}
The output of the SDPA module, $logits_{query}$, represents the classification logits of the query defect instance. The class with the highest logit value is identified as the predicted class.

    \subsection{Training Zip-Adapter-F Classifier}
    
Zip-Adapter-F enhances the visual features for better performance by fine-tuning the Zip-Adapter classifier, making both the ZIP module and the cached visual features of the support set learnable. 
Our Zip-Adapter-F combines a cache-based mechanism with an adapter-based mechanism, using the Zip-Adapter as the base model. The fine-tuning process involves two training objectives: (1) optimizing the cross-entropy (CE) loss between the predicted logits $logits_{query}$ and the labels $Y_{query}$. The CE loss $\mathcal{L}_{CE}$ is defined as:

    {\small\begin{equation}
    \mathcal{L}_{CE}(logits_{query}, Y_{query}) = -\sum_{i} y_{i} \log(p_{i}),
    \end{equation}}
    where $y_{i}$ and $p_{i}$ represents the label and predicted probability distribution for class $i$. \\
    The second part uses the triplet loss to optimize the intra-class compactness and inter-class separability of the adapted feature $f_{\text{adapted}}$ of the ZIP module within a batch. The triplet loss $\mathcal{L}_{triplet}$ is defined as:
    {\small\begin{equation}
        \begin{split}
    \mathcal{L}_{triplet}(F^{\prime}_{query}) &= \max ( d(F^{\prime}_{\text{anchor}}, F^{\prime}_{\text{positive}})  \\
    &- d(F^{\prime}_{\text{anchor}}, F^{\prime}_{\text{negative}}) + \alpha, 0 )
        \end{split},
    \end{equation}}
    where $F^{\prime}_{\text{anchor}}$, $F^{\prime}_{\text{positive}}$, and $F^{\prime}_{\text{negative}}$ are the embeddings (feature vectors) of an anchor sample, a positive sample (same class as an anchor), and a negative sample (different class from anchor), within a batch, respectively.  $d(\cdot, \cdot)$ is a distance function used to measure the similarity between embeddings. $\alpha$ is a margin hyperparameter that specifies the minimum difference between the distances of positive and negative pairs required for the loss to be zero.
    The overall loss function for finetuning Zip-Adapter-F is:

    {\small\begin{equation}
            \mathcal{L}_{Zip-Adapter-F} = \mathcal{L}_{CE} + \lambda \cdot \mathcal{L}_{triplet},
    \end{equation}}
    where $\lambda$ is a hyperparameter that balances the importance of the two parts in the overall loss.
    After Zip-Adapter-F is trained, it can be used to classify query defect instances similar to the Zip-Adapter classifier. 
        \begin{figure}[ht]
	\centering  % 更简洁的居中命令，不会增加额外的垂直空间
	\includegraphics[width=3.3in]{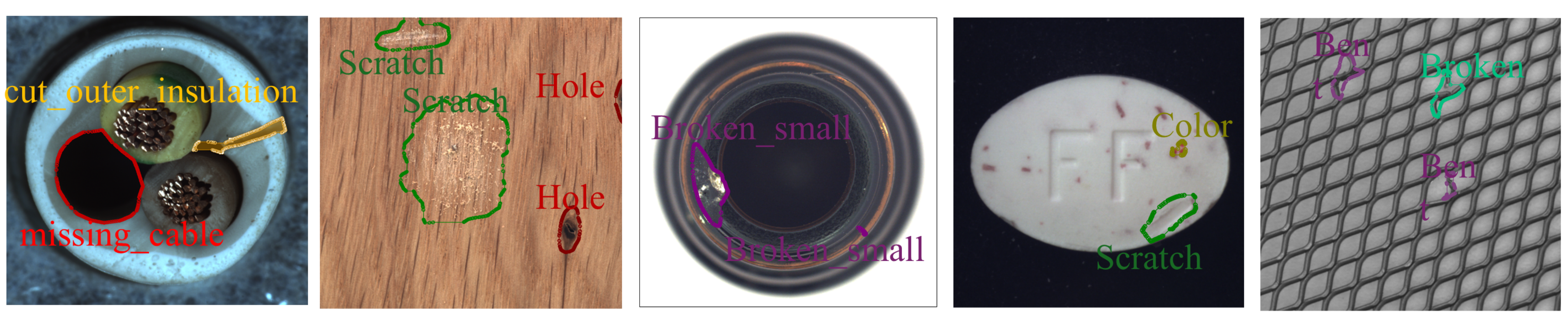}  % 确保图片文件名和路径正确

	\caption{ Some modified cases are displayed, in which there are multiple defect instances with different types. }  % 添加描述性的标题
	\label{fig:mvtecfs_sample}  % 为图像添加标签，便于文中引用
\end{figure}
    \begin{figure}[ht]
        \centering  % 更简洁的居中命令，不会增加额外的垂直空间
        \includegraphics[width=3.3in]{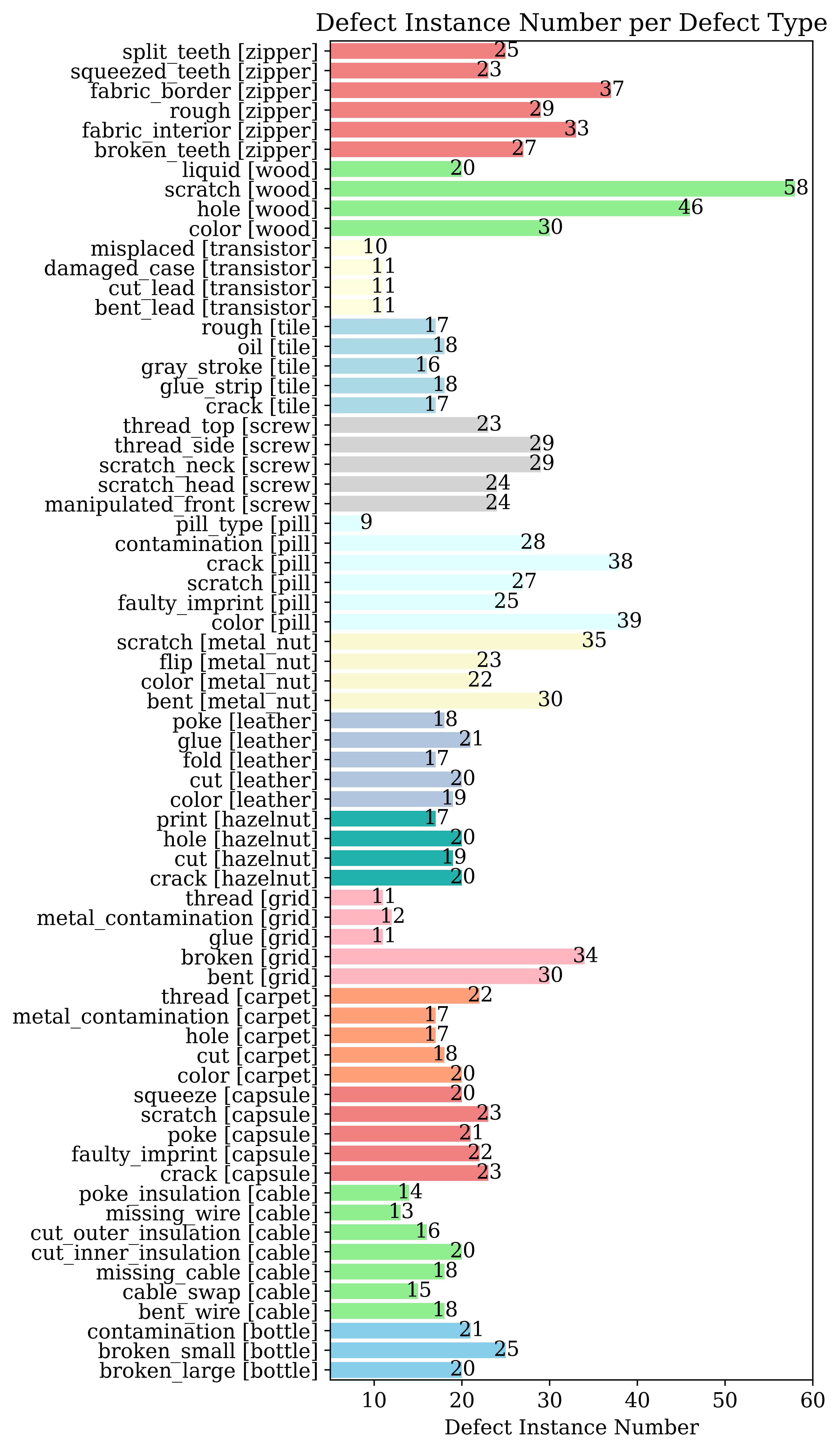}  % 确保图片文件名和路径正确
        \caption{Details of MVTec-FS dataset.}  % 添加描述性的标题
        \label{fig:mvtecfs}  % 为图像添加标签，便于文中引用
        
    \end{figure}

\section{MVTec-FS Dataset}

Although few-shot defect multi-classification has garnered considerable research attention, datasets like NEU-DET~\cite{neu_det} and MTD~\cite{mtd_dataset} are limited to a single product category. Recently, anomaly detection has also gained attention, with several multi-category datasets~\cite{MVTEC,mvtec_loco,visa_data} being proposed. However, these datasets are not designed for defect classification. The MVTec AD~\cite{MVTEC} dataset, the most popular benchmark for anomaly detection, features 15 product categories (5 textiles and 10 objects), offering significant diversity and generalization. In its original configuration, the training set consists of normal images, while the testing set includes both normal and defect images, with defect images labeled by masks. This dataset contains about 47 defect types (ranging from 8 to 26 images per type), making it suitable for FSDMC tasks. However, FSDMC tasks have rarely been studied on this dataset.
\\
We selected 1,228 defective images from the MVTec AD and labeled them with instance-level masks, creating a new benchmark dataset named MVTec-FS. Since the toothbrush category contains only one defect type, it was excluded from MVTec-FS. The number of defect instances per defect type is presented, totaling 46 types with instances ranging from 9 to 58, as shown in Fig~\ref*{fig:mvtecfs}. The original annotations were at the image level and did not account for multiple defects within a single image. We used the connected component algorithm to convert image-level masks to instance-level masks, followed by necessary human corrections. Some examples are shown in Figure~\ref{fig:mvtecfs_sample}. For each defect type, 50\% of the defects are used as the training set to sample the support set, while the remaining 50\% constitute the testing set (query set).

    \section{Experiments}

\subsection{Experiments Setting}

\begin{table*}[ht]
\centering
\setlength{\tabcolsep}{0.5mm} 
 {\fontsize{9}{9}\selectfont
\begin{tabular}{c|c|c|c|c|c|c|c|c|c|c|c|c|c|c|c|c}
    \toprule
    FS & Classifier & Carpet & Grid & Leather & Tile & Wood & Bottle & Cable & Capsule & Hazelnut & MetalNut & Pill & Screw & Transistor & Zipper & Average \\
    \midrule
    0 &  CLIP-ZeroShot & 25.0 & 51.0 & 43.2 & 42.9 & 75.7 & 34.4 & 21.1 & 20.8 & 62.9 & 36.0 & 14.8 & 23.3 & 33.3 & 39.0 & 37.4 \\
    \midrule
  
    \multirow{8}{*}{1}&  CLIP-Adapter  & 66.4 & 42.9 & 65.5 & 70.5 & 64.0 & 51.9 & 61.8 & 44.5 & 66.3 & 59.6 & 54.3 & 72.7 & 63.8 & 77.3 & 61.5\\
    & CLIP-ProtoNet  &  60.5 & 42.0 & 62.7 & 71.0 & 63.4 & 50.6 & 62.8 & 46.8 & 62.9 & 62.0 & 59.3 & 74.7 & 55.2 & 74.6 & 60.6 \\
    &  CLIP-KNN  & 59.6 & 42.0 & 62.7 & 69.5 & 62.9 & 50.6 & 62.8 & 47.2 & 62.9 & 61.6 & 58.5 & \textbf{75.0} & 55.2 & 74.4 & 60.4 \\
    &  CLIP-LinearProb  & 60.5 & 41.6 & 64.6 & 73.3 & 68.0 & 55.0 & 62.8 & 49.1 & 69.1 & 60.0 & 63.7 & 73.0 & 56.2 & 73.9 & 62.2 \\
    & Tip-Adapter  & 60.0 & 42.5 & 62.7 & 70.0 & 63.4 & 50.6 & 62.8 & 47.2 & 62.9 & 62.0 & 59.3 & 74.7 & 55.2 & 74.6 & 60.6 \\
    & Tip-Adapter-F  & 60.9 & 44.9 & 63.2 & 72.9 & 64.6 & 51.3 & 63.2 & 46.8 & 65.1 & 61.2 & 62.7 & 74.3 & 61.0 & 76.6 & 62.0 \\  \cmidrule(lr){2-17}
    &  \textbf{Zip-A (MVREC)} & \textbf{73.6} & \textbf{49.8} & \textbf{79.5} & \textbf{94.3} & \textbf{69.4} & \textbf{58.8} & \textbf{80.4} & \textbf{57.7} & \textbf{77.1} & \textbf{71.2} & \textbf{66.9} & 68.0 & \textbf{77.1} & \textbf{79.0} & \textbf{71.6} \\
    &  \textbf{Zip-A-F (MVREC)} & \textbf{78.6} & \textbf{50.6} & \textbf{82.3} & \textbf{96.2} & \textbf{71.4} & \textbf{58.1} & \textbf{77.5} & \textbf{60.4} & \textbf{77.1} & \textbf{71.6} & \textbf{72.4} & 67.3 & \textbf{89.5} & \textbf{79.3} & \textbf{73.7} \\
    \midrule

    \multirow{8}{*}{3}&  CLIP-Adapter  & 70.9 & 57.6 & 77.3 & 89.5 & 81.4 & 68.1 & 77.9 & 61.5 & 76.0 & 72.0 & 68.6 & 85.7 & 88.6 & 85.1 & 75.7 \\
    &  CLIP-ProtoNet  & 71.4 & 57.6 & 78.2 & 86.7 & 82.3 & 70.0 & 80.0 & 63.0 & 77.7 & 71.2 & 69.1 & 87.0 & 85.7 & 83.4 & 76.0  \\
    &  CLIP-KNN  & 71.4 & 57.6 & 78.2 & 86.7 & 82.3 & 70.0 & 80.0 & 63.0 & 77.7 & 71.2 & 69.1 & 87.0 & 85.7 & 83.4 & 76.0  \\
    &  CLIP-LinearProb  & 74.1 & 60.0 & 82.3 & 91.0 & 83.1 & 71.3 & 82.1 & 63.4 & 78.3 & 73.6 & 75.6 & 86.7 & 87.6 & 83.4 & 78.0 \\
    &  Tip-Adapter  & 65.5 & 55.9 & 77.3 & 86.2 & 75.7 & 68.1 & 79.0 & 60.4 & 77.7 & 73.2 & 67.4 & 86.3 & 84.8 & 74.9 & 73.7 \\
    &  Tip-Adapter-F  & 72.7 & 60.8 & 80.5 & 89.1 & 82.3 & 71.9 & 81.8 & 61.5 & 78.3 & 73.2 & 73.3 & 87.0 & 87.6 & 85.4 & 77.5 \\  \cmidrule(lr){2-17}
    &  \textbf{Zip-A (MVREC)} &  \textbf{81.8} & 56.7 & \textbf{87.7} & \textbf{97.1} & 82.9 & 63.1 & \textbf{91.6} & 63.4 & 76.6 & \textbf{78.4} & 74.8 & 80.0 & \textbf{97.1} & 82.9 & \textbf{79.6}  \\
    &  \textbf{Zip-A-F (MVREC)} & \textbf{85.0} & \textbf{71.8} & \textbf{90.9} & \textbf{97.6} & \textbf{90.0} & \textbf{76.9} & \textbf{93.0} & \textbf{74.0} & \textbf{82.3} & \textbf{83.6} & \textbf{83.5} & \textbf{88.3} & \textbf{100.0} & \textbf{88.8} & \textbf{86.1} \\
    \midrule

    \multirow{8}{*}{5}&  CLIP-Adapter  & 75.5 & 64.9 & 87.7 & 91.0 & 86.9 & 67.5 & 83.9 & 69.8 & 78.9 & 78.8 & 77.3 & 90.7 & 93.3 & 87.6 & 81.0 \\
    &  CLIP-ProtoNet  & 73.6 & 59.6 & 83.6 & 89.1 & 84.9 & 67.5 & 84.9 & 72.1 & 74.9 & 78.4 & 74.6 & 89.3 & \textbf{99.1} & 87.1 & 79.9 \\
    &  CLIP-KNN  & 74.1 & 55.5 & 81.8 & 89.5 & 79.7 & 65.6 & 78.3 & 62.3 & 73.1 & 79.2 & 67.6 & 89.7 & 97.1 & 79.8 & 76.7 \\
    &  CLIP-LinearProb  & 79.6 & 66.9 & 88.6 & 93.3 & 88.3 & 70.0 & 86.7 & 69.8 & 78.9 & 80.0 & 81.2 & \textbf{92.7} & 92.4 & 86.3 & 82.5 \\
    &  Tip-Adapter  & 72.7 & 59.2 & 80.9 & 88.1 & 84.9 & 63.8 & 82.1 & 65.7 & 77.7 & 78.4 & 67.6 & 90.3 & 98.1 & 78.3 & 77.7 \\
    &  Tip-Adapter-F  & 74.5 & 65.3 & 88.2 & 91.0 & 89.7 & 68.1 & 85.6 & 70.2 & 77.1 & 80.8 & 80.7 & 91.0 & 96.2 & 87.6 & 81.9 \\  \cmidrule(lr){2-17}
    &  \textbf{Zip-A (MVREC)} & \textbf{84.5} & 60.0 & \textbf{89.6} & \textbf{97.6} & 89.4 & 61.9 & \textbf{93.0} & \textbf{75.1} & \textbf{84.0} & \textbf{82.8} & 75.1 & 88.0 & \textbf{99.1} & 83.2 & \textbf{83.1}  \\
    &  \textbf{Zip-A-F (MVREC)}  & \textbf{85.9} & \textbf{80.8} & \textbf{92.7} & \textbf{97.6} & \textbf{96.6} & \textbf{77.5} & \textbf{93.0} & \textbf{81.1} & \textbf{88.6} & \textbf{91.2} & \textbf{84.7} & 92.0 & \textbf{100.0} & \textbf{90.0} & \textbf{89.4} \\
    \bottomrule
\end{tabular} 
}
\caption{Classification accuracy (\%) on MVTec-FS of different models.
AlphaCLIP is used to extract the visual feature for all classifiers. Zip-A and Zip-A-F stands for Zip-Adapter and  Zip-Adapter-F for short. The best results are highlighted in bold.}
\label{tab:classification_accuracy}
\end{table*}

\begin{table*}[ht]
    \centering

    \setlength{\tabcolsep}{1mm} 
    % \resizebox{0.9\linewidth}{!}{
     {\fontsize{9}{9}\selectfont
    \begin{tabular}{c|c|c|c|c|c|c|c|c|cc}
        \toprule
        FS & MVREC & CLIP-Adapter & CLIP-ProtoNet &  CLIP-KNN &   CLIP-LinearProb  &  Tip-Adapter &  Tip-Adapter-F & Zip-Adapter & Zip-Adapter-F \\
\midrule
\multirow{3}{*}{1} & \ding{55} & 61.5 & 60.6 & 60.4 & 62.2 & 60.6 & 62.0 & 60.6 & \textbf{62.4} \\
                   & \checkmark & 73.1 & 71.6 & 71.3 & 71.8 & 71.6 & 73.0 & 71.6 & \textbf{73.7} \\
                   & Gain & \textbf{+11.6} & +11.0 & +10.9 & +9.6 & +11.0 & +11 & +11.1 & +11.3 \\
\midrule
\multirow{3}{*}{3} & \ding{55} & 75.7 & 76.0 & 71.0 & \textbf{78.0} & 73.7 & 77.5 & 73.7 & 77.9 \\
                   & \checkmark & 85.7 & 83.4 & 78.9 & 84.8 & 79.6 & 85.7 & 79.6 & \textbf{86.1} \\
                   & Gain & \textbf{+10.0} & +7.6 & +7.9 & +6.8 & +5.9 & +8.2 & +5.9 & +8.2 \\
\midrule
\multirow{3}{*}{5} & \ding{55} & 81.0 & 79.9 & 76.7 & \textbf{82.5} & 77.7 & 81.9 & 77.7 & 82.2 \\
                   & \checkmark & 89.2 & 86.1 & 82.6 & 88.1 & 83.1 & 89.2 & 83.1 & \textbf{89.4} \\
                   & Gain & \textbf{+8.2} & +6.2 & +5.9 & +5.6 & +5.4 & +7.3 & +5.4 & +7.2 \\
        \bottomrule
    \end{tabular}
    }
    \caption{Gains of MVREC Representation and Zip-Adapter(-F) Classifiers across different Few-Shot Settings.}
    \label{tab:comparison_mvrec_zifa}
\end{table*}

\begin{table}[ht]
    \centering
    % \small % 调整字体大小为9pt
    \setlength{\tabcolsep}{1mm} 
     {\fontsize{9}{9}\selectfont
    \begin{tabular}{c|c|c|ccc}
        \toprule
        \multirow{2}{*}{Classifier} & \multirow{2}{*}{Crop Size}  & \multirow{2}{*}{Feature Extractor}  & \multicolumn{3}{c}{\textbf{Few Shot Setup}}  \\
        \cmidrule(l){4-6}
         & &   & \textbf{1} & \textbf{3} & \textbf{5}  \\ 
        \midrule
        \multirow{4}{*}{Zip-Adapter} & Defect & CLIP &  57.7 & 65.5 & 68.5 \\
        & Fixed  & CLIP&  58.1 & 64.8 & 68.6  \\
        & Fixed & AlphaCLIP  Wo.M   & 61.8 & 68.0 & 70.1  \\
        & Fixed & AlphaCLIP    & \textbf{71.6} & \textbf{79.6} & \textbf{83.1} \\ \midrule
        \multirow{4}{*}{Zip-Adapter-F}  & Defect &   CLIP   &  61.4 & 77.0 & 82.0  \\
        & Fixed &   CLIP    &  65.1 & 80.6 & 85.5  \\
        & Fixed & AlphaCLIP  Wo.M  &  66.3 & 79.6 & 85.4  \\
        & Fixed & AlphaCLIP    &   \textbf{73.7} & \textbf{86.1} & \textbf{89.4}  \\
        \bottomrule
    \end{tabular}
    }
    \caption{Classification accuracy (\%) on MVTec-FS with different region-context handling methods. Wo.M stands for Without Mask. Cropping by fixed size and using AlphaCLIP with a masked region-context both preserves and effectively utilizes the context.}

    \label{tab:region_context}
\end{table}

\begin{table}[!ht]
    \centering

    \setlength{\tabcolsep}{1mm} 
 
    % \scriptsize
    {\fontsize{9}{9}\selectfont
    \begin{tabular}{c|cccc|cccc}
        \toprule
        \multirow{2}{*}{Classifier}  & \multicolumn{4}{c|}{\textbf{Augmentation}}   & \multicolumn{3}{c}{\textbf{Few Shot Setup}}  \\
        \cmidrule(lr){2-8}
         & Scale & Rotate & Flip & Offset   & \textbf{1} & \textbf{3} & \textbf{5}  \\ 
        \midrule
        \multirow{7}{*}{Zip-Adapter}  & \ding{55} & \ding{55} & \ding{55} & \ding{55} & 60.6 & 73.7 & 77.7  \\
        & \checkmark & \ding{55} & \ding{55} & \ding{55} &    68.1 & 78.3 & 81.4  \\
        & \ding{55}   & \checkmark & \ding{55}  &\ding{55}  & 68.5 & 77.9 & 82.1 \\
        & \ding{55}    & \ding{55}  &\checkmark  &\ding{55}  &   55.3 & 64.8 & 68.2  \\ 
        &   \ding{55}  & \ding{55}  & \ding{55}  & \checkmark &   70.7 & 78.5 & 80.7 \\ 
        & \checkmark  & \checkmark &\ding{55}  &\ding{55}  &   70.5 & 78.4 & 82.0  \\
        & \checkmark   & \ding{55} &\checkmark  &\ding{55}  &   60.3 & 71.1 & 74.5  \\ 
        &  \checkmark  &\ding{55}  & \ding{55}  & \checkmark & \textbf{71.6} & \textbf{79.6} & \textbf{83.1} \\ \midrule
        \multirow{7}{*}{Zip-Adapter-F} & \ding{55} & \ding{55} & \ding{55} & \ding{55} & 62.4 & 77.9 & 82.2 \\
        & \checkmark & \ding{55} &\ding{55}  &\ding{55}   &  69.3 & 82.9 & 86.4   \\
        &  \ding{55}  & \checkmark & \ding{55}  & \ding{55}  & 70.5 & 83.2 & 86.8    \\
        &  \ding{55}   & \ding{55}  &\checkmark  & \ding{55} & 54.7 & 72.6 & 78.3  \\ 
        &  \ding{55}   &\ding{55}  & \ding{55}  & \checkmark &    72.4 & 83.8 & 87.1  \\ 
        & \checkmark  & \checkmark &\ding{55}  &\ding{55}  &   73.1 & 84.3 & 88.4  \\
        & \checkmark   &\ding{55}  &\checkmark  & \ding{55}  &   62.6 & 79.5 & 85.0   \\ 
        &  \checkmark  & \ding{55}  & \ding{55}  & \checkmark &   \textbf{73.7} & \textbf{86.1} & \textbf{89.4}  \\
        \bottomrule
    \end{tabular}
    }
    \caption{Classification accuracy (\%) on MVTec-FS of different augmentation methods.}
    \label{tab:context_augment}
\end{table}

    \begin{table}[!ht]
        \centering
        % \small % 调整字体大小为9pt
        \setlength{\tabcolsep}{1mm} 
     
        % \scriptsize
      {\fontsize{9}{9}\selectfont
        \begin{tabular}{cc|ccc}
            \toprule
             \multicolumn{2}{c|}{\textbf{Zip-Adapter-F Config}}   & \multicolumn{3}{c}{\textbf{Few Shot Setup}}  \\
             \midrule
              Trainable support feature & Trainable ZIP      & \textbf{1} & \textbf{3} & \textbf{5}  \\ 
             \midrule
            %  \ding{55}  & \ding{55} &  73.77 & 86.15 & 89.45 \\
             \ding{51}  & \ding{55} &  72.67 & 85.55 & 89.39\\
             \ding{55}  & \ding{51} &  73.45 & 85.89 & 89.32 \\
             \ding{51}  & \ding{51} &  \textbf{73.74 }& \textbf{86.12} & \textbf{89.41 }\\
            \bottomrule
    
        \end{tabular}
        }
        \caption{Classification accuracy (\%) on MVTec-FS of different training setttings of ZIFA-Adapter-F.}
        \label{tab:zifa_setting}
    \end{table}

We conducted experiments on the MVTec-FS dataset and four other datasets to evaluate our MVREC using accuracy metrics. The few-shot setup is defined as N-way K-shot, where K is set to 1, 3, or 5. The support set is sampled from the training set, and the query set consists of all images in the testing set. The evaluation was conducted on the query set for each of the five sampled support sets, and the average classification accuracy was calculated to provide a more robust assessment. Ablation studies were performed on the MVTec-FS dataset to assess the effectiveness of the various components. For AlphaCLIP, we selected the ViT-L/14~\cite{vit} backbone. The MVREC visual feature used three scales, representing the commonly used large, medium, and small settings. The number of offsets was set to 9, based on a grid layout similar to a tic-tac-toe board. We set $\beta$ to 32 for the Zip-Adapter and 1 for the Zip-Adapter-F classifiers. When training the Zip-Adapter-F, we used the AdamW optimizer with a learning rate of 0.0001. The model was updated for 500 iterations, training on all MVREC features of the support set in each iteration. For the triplet loss item, the hyperparameters $\alpha$ and $\lambda$ were set to 0.5 and 4, respectively.

In our experiment, we evaluated a variety of baseline classifiers based on the AlphaCLIP backbone, including: 1. \textbf{CLIP-ZeroShot}~\cite{clip}: This approach leverages the zero-shot capability of the CLIP model. We generated text embeddings for each class description and computed the similarity between the test image embeddings and these text embeddings, classifying them based on the highest similarity. 2. \textbf{CLIP-KNN}: This method uses the K-Nearest Neighbors algorithm with the CLIP features. The most similar K (K=1) support samples are retrieved for a query sample, and the class with the majority vote is selected as the prediction. 3. \textbf{CLIP-ProtoNet}: This approach builds a Prototypical Network~\cite{proto_net} on top of the CLIP, where proxy features representing each class are calculated from the support set, and test images are classified based on their similarity to these class proxies. 4. \textbf{CLIP-Adapter}~\cite{clip_adapter}: This method involves adding adapter layers on top of the CLIP. These layers are trained for the new classification task, adjusting the image features accordingly. 5. \textbf{Tip-Adapter}~\cite{tip_adpater}: This method constructs a key-value cache model using CLIP-extracted features from the few-shot data and performs recognition in a retrieval-based manner. Tip-Adapter-F treats the visual cache as learnable parameters and optimizes them to improve performance.

\subsection{Results on MVTec-FS Dataset}

In Table~\ref{tab:classification_accuracy}, we present the classification accuracy (\%) of various few-shot models evaluated on the MVTec-FS dataset under different few-shot learning configurations. To ensure a fair comparison, we report results across several few-shot settings, specifically with 0, 1, 3, and 5 shots. The accuracies for 14 product categories, as well as the average accuracy, are reported in separate columns. As shown in the table, our MVREC demonstrates outstanding performance in all few-shot setups, regardless of whether Zip-Adapter or Zip-Adapter-F is used. The Zip-Adapter-F achieves the highest accuracy of 89.4\% with 5 shots, which is 6.9\% higher than the second-best, LinearProb. By comparing different product categories, it is evident that the Zip-Adapter-F achieves the highest accuracy in most categories, demonstrating its effectiveness in few-shot learning scenarios.

    \subsection{Ablation Study}
    \subsubsection{Contributions of MVREC feature and Zip-Adapter(-F).} 
    First, we assess the effectiveness of the MVREC feature and Zip-Adapter(-F) classifiers by comparing their performance to other classifiers, regardless of MVREC usage. As shown in Table~\ref{tab:comparison_mvrec_zifa}, the MVREC feature consistently boosts the performance of all classifiers across different few-shot settings, with the most significant gain of 11.6\% in CLIP-Adapter with 1-shot, demonstrating its general effectiveness for few-shot defect classification. Zip-Adapter-F consistently outperforms most classifiers, regardless of MVREC use, highlighting its inherent strength. When combined with MVREC, Zip-Adapter-F achieves the best results across all few-shot settings, maximizing its potential and making it the most effective approach for FSDMC. Notably, Zip-Adapter and Tip-Adapter yield identical results before training, as they are mathematically equivalent at that stage.
    \subsubsection{Mask Region-Context.}
    We investigate the impact of the mask region-context on model performance. As previously mentioned, the mask region-context helps the model focus on the defect instance without cropping the region based on defect size, which could otherwise result in the loss of contextual information. We evaluate two scenarios where the mask region-context is removed: (1) using a whole-foreground mask as the region-context input to AlphaCLIP, and (2) using CLIP without any mask region-context. The results, shown in Table~\ref{tab:region_context}, demonstrate that removing the mask context leads to a noticeable decrease in accuracy. We also consider the impact of different cropping styles. When cropping by defect size and using vanilla CLIP, the worst results are obtained, further emphasizing the importance of mask region-context. Cropping by fixed size and using AlphaCLIP as the feature extractor achieves the best performance, highlighting the effectiveness of MVREC.
    \subsubsection{Multi-View Context Augmentation.}  
    From the results in Table~\ref{tab:region_context}, we observe that different augmentation methods have varying impacts on classification accuracy. When single augmentations are used, multi-scale, multi-offset, and multi-rotation augmentations show significant improvements in both Zip-Adapter and Zip-Adapter-F. When double augmentations are applied, the combination of multi-scale and multi-offset yields the best results, indicating that these augmentations are complementary and can be combined to achieve better performance. Multi-scale augmentation allows the model to learn features at various resolutions, which is crucial for capturing both fine and coarse details in the images. Meanwhile, multi-offset augmentation helps the model learn robust features by shifting the image and mask context, thereby improving the model's robustness.% and accuracy in real-world applications.

\begin{table}[!ht]
\centering
\setlength{\tabcolsep}{0.8mm}
\fontsize{9}{9}\selectfont
{
\begin{tabular}{c|c|c}
\toprule
\textbf{Dataset} & \textbf{Defect Type (Instance Number)} & \textbf{Annotations} \\ \midrule
\multirow{3}[0]{*}{NEU\_DET}
 & Crazing(689), Pitted\_surface(432),  & \multirow{3}[0]{*}{Bbox} \\ 
 & Rolled\_in\_scale(628), Patches(881),\\
 & Scratches(548), Inclusion(1011) \\ 
\midrule
\multirow{3}[0]{*}{PCB}
 & Spurious\_copper(503), Short(491),  & \multirow{3}[0]{*}{Bbox} \\ & Spur(488), Mouse\_bite(492),  \\ & Missing\_hole(497), Open\_circuit(482) \\
\midrule
\multirow{2}[0]{*}{MTD}
 & Break(108), Crack(69), Fray(37), & \multirow{2}[0]{*}{Mask}\\
 &Uneven(103), Blowhole(115) \\
\midrule
\multirow{4}[0]{*}{AITEX}
 & Broken\_end(11), Broken\_yarn(16), & \multirow{4}[0]{*}{Mask} \\
 &Cuts\_elvage(12), Weftcrack(15), \\
   &Fuzzyball(42), Nep(19),\\
  & Broken\_pick (65), \\
\hline
\end{tabular}
}
\caption{Defect types and their counts in Other four datasets.}
\label{table:other_four_dataset}
\end{table}

    \subsubsection{Different Training Setting of Zip-Adapter(-F).}
    As shown in Figure~\ref{tab:zifa_setting}, the combination of trainable support features and a trainable ZIP module resulted in the highest accuracy across all few-shot setups.

    \subsection{Visualization}

\begin{figure}[!ht]
    \centering  % 更简洁的居中命令，不会增加额外的垂直空间
    \includegraphics[width=3.3in]{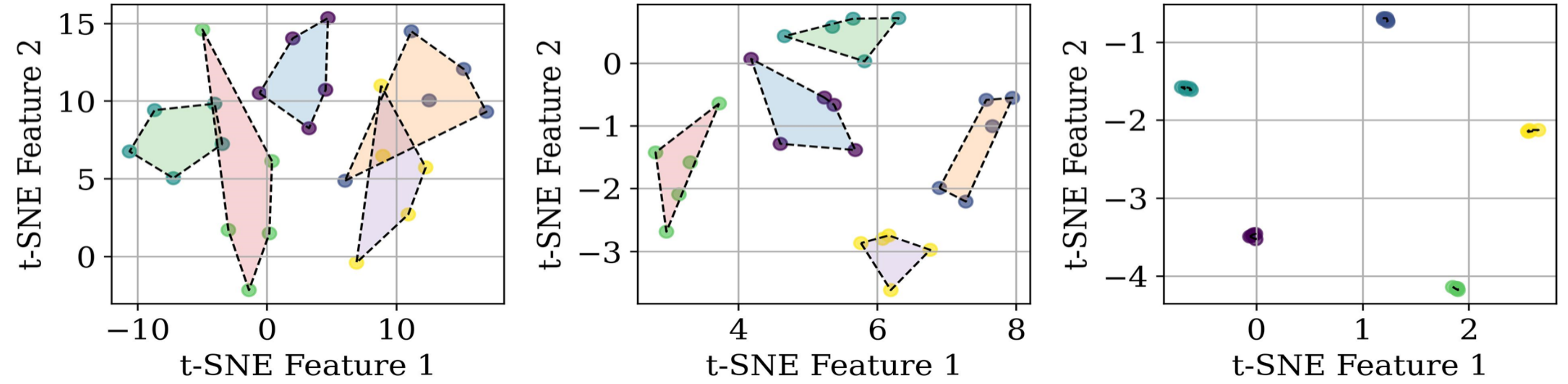}  % 确保图片文件名和路径正确
    \caption{t-SNE projections of the MVREC features  $F_{SUPP}$ for support set. From left to right are 1)  $F_{SUPP}$ without multi-view augmentation. 2)  $F_{SUPP}$ 3)  finetuned $F_{SUPP}$. }  % 添加描述性的标题
    \label{fig:tsne}  % 为图像添加标签，便于文中引用
\end{figure}

    To better illustrate the function of MVREC, we used t-SNE~\cite{tsne} to visualize the support MVREC features in Zip-Adapter-F, as shown in Figure~\ref{fig:tsne}. Different colors represent 5 defect classes from the 5-shot leather images of the MVTec-FS. The changes in the distribution indicate that multi-view augmentation and fine-tuning help the model learn more discriminative features.

    \subsection{Comparison on Other Datasets}

       \begin{figure}[!ht]
    \centering  % 更简洁的居中命令，不会增加额外的垂直空间
    \includegraphics[width=3.3in]{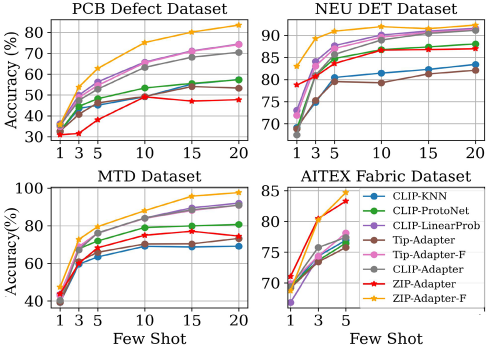}  % 确保图片文件名和路径正确
    \caption{Classification accuracy (\%) on other four datasets of different models with quantitative values.}  % 添加描述性的标题
    \label{fig:other_data}  % 为图像添加标签，便于文中引用
\end{figure}
We evaluated MVREC on several public datasets, including: \textbf{1) NEU-DET}~\cite{neu_det}, a metal surface defect dataset for detection model research; \textbf{2) PCB Defect Dataset}~\cite{huang2019pcb}, released by The Open Lab on Human-Robot Interaction of Peking University; \textbf{3) Magnetic Tile Surface Defects (MTD)}~\cite{mtd_dataset}, which contains 6 common magnetic tile defects; and \textbf{4) AITEX Fabric Defect}~\cite{aitex_fabric}, a fabric defect dataset with 12 types of defects, from which seven defect types with at least 10 samples are selected. For each dataset, 50\% of the data is used as the training set for sampling the support set, and the other 50\% is used as the testing set (query set). In addition to 1, 3, and 5 shots, we also evaluated performance with 10, 15, and 20 shots on the NEU-DET, PCB Defect, and MTD datasets for a more comprehensive comparison. The results, in Figure~\ref{fig:other_data}, demonstrate that Zip-Adapter-F (MVREC) achieves the best performance on all datasets, with performance improving as the number of shots increases.% Overall, these results demonstrate the generalization capability of our method across different datasets with various types of annotations.

\section{Discussion and Conclusion}
This paper introduces MVREC, an instance-level few-shot classification approach that can be applied to various labeled formats, such as bounding boxes and masks. Extensive experiments on five datasets demonstrate that it is a versatile and effective approach for FSDMC.\\
\textbf{Limitations.} First, we have not yet explored the unified model that can be developed to handle different defect datasets after a single training session. Second, our study mainly focuses on using image features extracted by CLIP, without exploring the potential of CLIP's text encoder for multi-model research. We hope this work inspires future research and the development of more advanced methods.

\section*{Acknowledgements}
This research is supported by Laboratory for
Artificial Intelligence in Design (Project Code: RP3-3) under InnoHK
Research Clusters, Hong Kong SAR Government.

\bibliography{MVREC}

\clearpage
\section*{Appendix}
\subsection{More details of MVTec-FS}

\begin{figure}[!ht]
    \centering
    \includegraphics[width=3in]{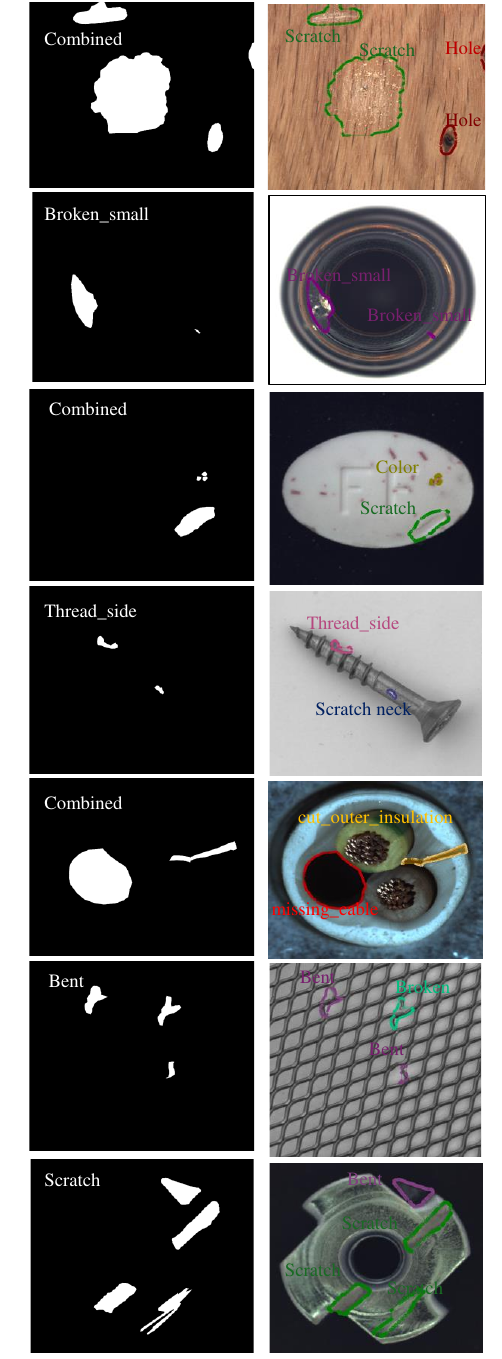}  
    \caption{Some examples of these label modifications in MVTec-FS. On the left are the image-level segmentation annotations from MVTec AD, while on the right are the instance-level segmentation annotations from MVTec-FS.}
    \label{fig:com_sample}
\end{figure}

 In creating the MVTec-FS dataset, we began by selecting all 1,228 anomaly images and their corresponding masks from the testing set of the MVTec AD dataset. The original anomaly masks were annotated at the image level, which is a coarse form of labeling. For instance, as shown in Figure \ref{fig:com_sample}, when multiple different types of anomalies appear in the same image, they are labeled as a single "combined" type. Additionally, when multiple defects of the same type appear in an image, the image-level mask is treated as a single instance. Given these problems, we refined the mask labels by converting them into instance-level defect masks using the connected component algorithm, assigning a class label to each defect instance.  Subsequently, we manually reviewed and adjusted the defect instance labels to ensure accurate labeling of each defect instance. Figure \ref{fig:com_sample} illustrates examples of these label modifications. Specifically, the label modification involved two main actions: 1) verifying and revising the defect instance masks, and 2) checking and correcting the instance class labels. For the "combined" type, where multiple defect instances exist within a single image, we modified the instance class labels to ensure that each defect instance was labeled correctly. \\
The MVTec-FS dataset, as summarized in Table \ref{tab:mvtec_fs}, showcases a diverse collection of sub-datasets, each representing different product categories with varying numbers of anomaly types and defect instances. Across these sub-datasets, the number of anomaly categories ranges from 3 to 7, reflecting the distinct defect characteristics of each product type. Notably, each sub-dataset within MVTec-FS is carefully constructed to ensure that there are at least five instances of each anomaly type in the training set. This design allows for effective few-shot learning experiments, supporting scenarios where 1-shot, 3-shot, and 5-shot learning paradigms can be evaluated. \\
Moreover, MVTec-FS is not only suitable for few-shot classification tasks but also serves as a valuable resource for \textbf{few-shot object detection} and\textbf{ unified multi-modal classification} tasks. We hope that this dataset will stimulate further research in these areas, fostering advancements in both few-shot learning and multi-modal learning fields.

% \begin{figure*}[ht]
%     \centering  % 更简洁的居中命令，不会增加额外的垂直空间
%     \includegraphics[width=4in]{LaTeX/fs_pic/com_sample.pdf}  
%     \caption{Some examples of these label modifications in MVTec-FS. }  % 添加描述性的标题
%     \label{fig:com_sample}  % 为图像添加标签，便于文中引用
% \end{figure*}

\begin{table*}[t]
\centering
\renewcommand{\arraystretch}{0.5}  % 压缩行距
% \resizebox{0.95\linewidth}{!}{
\setlength{\tabcolsep}{5pt} % 调整列间距
\begin{tabular}{l|rr|r|l|rr}
\toprule
\multirow{2}{*}{Sub\_dataset} & \multicolumn{2}{c|}{Image} & \multirow{2}{*}{Anomaly Type Num } & \multirow{2}{*}{Anomaly Type} & \multicolumn{2}{c}{Instance Num} \\ \cmidrule(lr){2-3} \cmidrule(lr){6-7}
 & Training & Testing & & & Training & Testing \\
\midrule
\multirow{3}{*}{bottle} & \multirow{3}{*}{32} & \multirow{3}{*}{31} & \multirow{3}{*}{3} & broken\_large & 10 & 10 \\
 &  &  &  & broken\_small & 13 & 12 \\
 &  &  &  & contamination & 11 & 10 \\ \hline
\multirow{7}{*}{cable} & \multirow{7}{*}{47} & \multirow{7}{*}{45} & \multirow{7}{*}{7} & bent\_wire & 9 & 9 \\
 &  &  &  & cable\_swap & 7 & 8 \\
 &  &  &  & missing\_cable & 7 & 11 \\
 &  &  &  & cut\_inner\_insulation & 9 & 11 \\
 &  &  &  & cut\_outer\_insulation & 8 & 8 \\
 &  &  &  & missing\_wire & 8 & 5 \\
 &  &  &  & poke\_insulation & 9 & 5 \\ \hline
\multirow{5}{*}{capsule} & \multirow{5}{*}{56} & \multirow{5}{*}{53} & \multirow{5}{*}{5} & crack & 12 & 11 \\
 &  &  &  & faulty\_imprint & 11 & 11 \\
 &  &  &  & poke & 11 & 10 \\
 &  &  &  & scratch & 12 & 11 \\
 &  &  &  & squeeze & 10 & 10 \\ \hline
\multirow{5}{*}{carpet} & \multirow{5}{*}{47} & \multirow{5}{*}{42} & \multirow{5}{*}{5} & color & 11 & 9 \\
 &  &  &  & cut & 9 & 9 \\
 &  &  &  & hole & 9 & 8 \\
 &  &  &  & metal\_contamination & 9 & 8 \\
 &  &  &  & thread & 12 & 10 \\ \hline
\multirow{5}{*}{grid} & \multirow{5}{*}{30} & \multirow{5}{*}{27} & \multirow{5}{*}{5} & bent & 14 & 16 \\
 &  &  &  & broken & 16 & 18 \\
 &  &  &  & glue & 6 & 5 \\
 &  &  &  & metal\_contamination & 7 & 5 \\
 &  &  &  & thread & 6 & 5 \\ \hline
\multirow{4}{*}{hazelnut} & \multirow{4}{*}{36} & \multirow{4}{*}{34} & \multirow{4}{*}{4} & crack & 11 & 9 \\
 &  &  &  & cut & 11 & 8 \\
 &  &  &  & hole & 10 & 10 \\
 &  &  &  & print & 9 & 8 \\ \hline
\multirow{5}{*}{leather} & \multirow{5}{*}{48} & \multirow{5}{*}{44} & \multirow{5}{*}{5} & color & 10 & 9 \\
 &  &  &  & cut & 11 & 9 \\
 &  &  &  & fold & 9 & 8 \\
 &  &  &  & glue & 12 & 9 \\
 &  &  &  & poke & 9 & 9 \\ \hline
\multirow{4}{*}{metal\_nut} & \multirow{4}{*}{48} & \multirow{4}{*}{45} & \multirow{4}{*}{4} & bent & 18 & 12 \\
 &  &  &  & color & 11 & 11 \\
 &  &  &  & flip & 12 & 11 \\
 &  &  &  & scratch & 19 & 16 \\ \hline
\multirow{6}{*}{pill} & \multirow{6}{*}{73} & \multirow{6}{*}{68} & \multirow{6}{*}{6} & color & 20 & 19 \\
 &  &  &  & faulty\_imprint & 13 & 12 \\
 &  &  &  & scratch & 13 & 14 \\
 &  &  &  & crack & 18 & 20 \\
 &  &  &  & contamination & 16 & 12 \\
 &  &  &  & pill\_type & 5 & 4 \\ \hline
\multirow{5}{*}{screw} & \multirow{5}{*}{61} & \multirow{5}{*}{58} & \multirow{5}{*}{5} & manipulated\_front & 12 & 12 \\
 &  &  &  & scratch\_head & 12 & 12 \\
 &  &  &  & scratch\_neck & 16 & 13 \\
 &  &  &  & thread\_side & 17 & 12 \\
 &  &  &  & thread\_top & 12 & 11 \\ \hline
\multirow{5}{*}{tile} & \multirow{5}{*}{43} & \multirow{5}{*}{41} & \multirow{5}{*}{5} & crack & 9 & 8 \\
 &  &  &  & glue\_strip & 9 & 9 \\
 &  &  &  & gray\_stroke & 8 & 8 \\
 &  &  &  & oil & 9 & 9 \\
 &  &  &  & rough & 9 & 8 \\ \hline
\multirow{4}{*}{transistor} & \multirow{4}{*}{20} & \multirow{4}{*}{20} & \multirow{4}{*}{4} & bent\_lead & 5 & 6 \\
 &  &  &  & cut\_lead & 6 & 5 \\
 &  &  &  & damaged\_case & 6 & 5 \\
 &  &  &  & misplaced & 5 & 5 \\ \hline
\multirow{4}{*}{wood} & \multirow{4}{*}{32} & \multirow{4}{*}{28} & \multirow{4}{*}{4} & color & 18 & 12 \\
 &  &  &  & hole & 26 & 20 \\
 &  &  &  & scratch & 28 & 30 \\
 &  &  &  & liquid & 12 & 8 \\ \hline
\multirow{6}{*}{zipper} & \multirow{6}{*}{61} & \multirow{6}{*}{58} & \multirow{6}{*}{6} & broken\_teeth & 16 & 11 \\
 &  &  &  & fabric\_interior & 14 & 19 \\
 &  &  &  & rough & 17 & 12 \\
 &  &  &  & fabric\_border & 19 & 18 \\
 &  &  &  & squeezed\_teeth & 12 & 11 \\
 &  &  &  & split\_teeth & 14 & 11 \\
\bottomrule
\end{tabular}
% }
\caption{MVTec FS dataset details. Training set is for sampling to support set and testing set are used as query set for evaluation.} 
\label{tab:mvtec_fs}
\end{table*}

\end{document}